# Exact and Approximate Inference in Associative Hierarchical Networks using Graph Cuts

**Chris Russell**[1] and **Ľubor Ladický** [1]      **Pushmeet Kohli**[2]                **Philip H.S. Torr**[1]
Oxford Brookes University      Microsoft Research Cambridge      Oxford Brookes University
[1] http://cms.brookes.ac.uk/research/visiongroup/         [2] http://research.microsoft.com/en-us/um/people/pkohli/

## Abstract

Markov Networks are widely used through out computer vision and machine learning. An important subclass are the Associative Markov Networks which are used in a wide variety of applications. For these networks a good approximate minimum cost solution can be found efficiently using graph cut based move making algorithms such as *alpha*-expansion. Recently a related model has been proposed, the associative hierarchical network, which provides a natural generalisation of the Associative Markov Network for higher order cliques (i.e. clique size greater than two). This method provides a good model for object class segmentation problem in computer vision.

Within this paper we briefly describe the associative hierarchical network and provide a computationally efficient method for approximate inference based on graph cuts. Our method performs well for networks containing hundreds of thousand of variables, and higher order potentials are defined over cliques containing tens of thousands of variables. Due to the size of these problems standard linear programming techniques are inapplicable. We show that our method has a bound of 4 for the solution of general associative hierarchical network with arbitrary clique size noting that few results on bounds exist for the solution of labelling of Markov Networks with higher order cliques.

## 1 INTRODUCTION

The last few decades have seen the emergence of Markov networks or random fields as the most widely used probabilistic model for formulating problems in machine learning and computer vision. This interest has led to a large amount of work on the problem of estimating the maximum a posteriori (MAP) solution of a random field (Szeliski et al., 2006). However, most of this research effort has focused on inference over pairwise Markov networks. Of particular interest are the families of associative pairwise potentials (Taskar et al., 2004), in which connected variables are assumed to be more likely than not to share the same label. Inference algorithms targeting these associative potentials, which include truncated convex costs (Kumar and Torr, 2008), metrics (Boykov et al., 2001), and semi metrics (Kumar and Koller, 2009), often carry bounds which guarantee the cost of the solution found must lie within a bound, specified as a fixed factor of $n$ of the cost of the minimal solution.

Although higher order Markov networks (i.e. those with a clique size greater than two) have been used to obtain impressive results for a number of challenging problems in computer vision (Roth and Black, 2005; Komodakis and Paragios, 2009; Vicente et al., 2009; Ladicky et al., 2010), the problem of bounded higher order inference has been largely ignored.

In this paper, we address the problem of performing graph cut based inference in a new model: the Associative Hierarchical Networks (AHNs) (Ladicky et al., 2009), which includes the higher order Associative Markov Networks (AMNs) (Taskar et al., 2004) or $P^n$ potentials (Kohli et al., 2007) and the Robust $P^n$ (Kohli et al., 2008) model as special cases, and derive a bound of 4.

This family of AHNs have been successfully applied to diverse problems such as object class recognition, document classification and texture based video segmentation, where they obtain state of the art results. Note that in our earlier work Ladicky et al. (2009), the problem of inference is not discussed at all; it shows how these hierarchical models can be used for scene understanding, and how learning is possible under the assumption that the model is tractable.

For a set of variables $\mathbf{x}^{(1)}$ AHNs are characterised by energies (or costs) of the form:

$$E(\mathbf{x}^{(1)}) = E'(\mathbf{x}^{(1)}) + \min_{\mathbf{x}^a} E^a(\mathbf{x}^{(1)}, \mathbf{x}^a) \qquad (1)$$

where $E'$ and $E^a$ are pairwise AMNs and $\mathbf{x}^a$ is a set of auxiliary variables. The AHN is a AMN containing higher order cliques, defined only in terms of $\mathbf{x}^{(1)}$, but can also be seen as a pairwise AMN defined in terms of $\mathbf{x}^{(1)}$ and $\mathbf{x}^a$. We propose new move making algorithms over the pairwise energy $E'(\mathbf{x}^{(1)}) + E^a(\mathbf{x}^{(1)}, \mathbf{x}^a)$ which have the important property of *transformational optimality*.

Move making algorithms function by efficiently searching through a set of candidate labellings and proposing the *optimal* candidate i.e. the one with the lowest energy to move to. The set of candidates is then updated, and the algorithm repeats till convergence.

We call a move making algorithm *transformationally optimal* if and only if any proposed move $(\mathbf{x}^*, \mathbf{x}^a)$ satisfies the property:

$$E(\mathbf{x}^*) = E'(\mathbf{x}^*) + E^a(\mathbf{x}^*, \mathbf{x}^a). \qquad (2)$$

Experimentally, our transformationally optimal algorithms converge faster, and to better solutions than standard approaches, such as $\alpha$-expansion. Moreover, unlike standard approaches, our transformationally optimal algorithms always find the exact solution for binary AHNs.

**Outline of the paper** In section 2 we introduce the notation used in the rest of the paper. Existing models generalised by the associative hierarchical network, and the full definition of AHNs are given in section 3. In section 4 we discuss work on efficient inference, and show how the pairwise form of associative hierarchical networks can be minimised using the $\alpha$-expansion algorithm, and derive bounds for our approach. Section 5 discusses the application of novel move making algorithms to such energies, and we show that under our formulation the moves of the robust $P^n$ model become equivalent to a more general form of range moves over unordered sets. We derive transformational optimality results over hierarchies of these potentials, guaranteeing the optimality of the moves proposed. We experimentally verify the effectiveness of our approach against other methods in section 6, and conclude in section 7.

## 2 NOTATION

Consider an AMN defined over a set of latent variables $\mathbf{x} = \{x_i | i \in \mathcal{V}\}$ where $\mathcal{V} = \{1, 2, ..., n\}$. Each random variable $x_i$ can take a label from the label set $\mathcal{L} = \{l_1, l_2, ..., l_k\}$. Let $\mathcal{C}$ represent a set of subsets of $\mathcal{V}$ (i.e., cliques), over which the AMN is defined. The MAP solution of a AMN can be found by minimising an energy function $E : \mathcal{L}^n \to \mathbb{R}$. These energy functions can typically be written as a sum of potential functions: $E(\mathbf{x}) = \sum_{c \in \mathcal{C}} \psi_c(\mathbf{x_c})$, where $\mathbf{x_c}$ represents the set of variables included in any clique $c \in \mathcal{C}$. We refer to functions defined over cliques of size one as unary potentials and denote them by $\psi_i : \mathcal{L} \to \mathbb{R}$ where the subscript $i$ denotes the index of the variable over which the potential is defined. Similarly, functions defined over cliques of size two are referred to as pairwise potentials and denoted: $\psi_{ij} : \mathcal{L}^2 \to \mathbb{R}$.

Potentials defined over cliques of size greater than two i.e. $\psi_c : \mathcal{L}^{|c|} \to \mathbb{R}, |c| > 2$ will be called higher order potentials, where $|c|$ represents the number of variables included in the clique (also called the clique order). We will call an energy function pairwise if it contains no potentials defined over cliques of size greater than 2.

At points within the paper, we will want to distinguish between the original variables of the energy function, whose optimal values we are attempting to find, and the auxiliary variables which we will introduce to convert our higher order function into a pairwise one. We refer to the original variables as the base layer $\mathbf{x}^{(1)}$ (as they lie at the bottom of the hierarchical network). All auxiliary variables at any level $h$ of the hierarchy are denoted by $\mathbf{x}^{(h)}$. The set of indices of variables constituting level $h$ of the hierarchy is denote by $\mathcal{V}^h$. Similarly, the set of all pairwise interactions at level $h$ is denoted by $\mathcal{E}^h$.

## 3 ASSOCIATIVE HIERARCHICAL NETWORKS

**Existing higher-order models** Taskar et al. (2004) proposed the use of higher order potentials that encourage the entirety of a clique to take some label, and discusses how they can be applied to predicting protein interactions and document classification. These potentials were introduced into computer vision along with an efficient graph cut based method of inference, as the strict $P^n$ Potts model (Kohli et al., 2007).

A generalisation of this approach was proposed by Kohli et al. (2008), who observed that in the image labelling problem, most (but not all) pixels belonging to image segments computed using an unsupervised clustering/segmentation algorithm take the same object label. They proposed a higher order MRF over segment based cliques. The energy took the form:

$$E(\mathbf{x}) = \sum_{i \in \mathcal{V}} \psi_i(x_i) + \sum_{ij \in \mathcal{E}} \psi_{ij}(x_i, x_j) + \sum_{c \in \mathcal{C}} \psi_c(\mathbf{x}_c), \quad (3)$$

$$\text{where } \psi_c(\mathbf{x}_c) = \min_{l \in \mathcal{L}} \left( \gamma_c^{max}, \gamma_c^l + \sum_{i \in c} k_c^i \Delta(x_i \neq l) \right). \quad (4)$$

The potential function parameters $k_c^i$, $\gamma_c^l$, and $\gamma_c^{max}$ are subject to the restriction that $k_c^i \geq 0$ and $\gamma_c^l \leq \gamma_c^{max}, \forall l \in \mathcal{L}$. $\Delta$ denotes the Kronecker delta, an indicator function taking a value of 1 if the statement following it is true and 0 if false.

These potentials can be understood as a truncated majority voting scheme on the base layer. Where possible, they encourage the entirety of the clique to assume one consistent labelling. However, beyond a certain threshold of disagreement they implicitly recognise that no consistent labelling is likely to occur, and no further penalty is paid for increasing heterogeneity.

We now demonstrate that the higher order potentials $\psi_c(\mathbf{x}_c)$ of the Robust $P^n$ model (4) can be represented by an equivalent pairwise function $\psi_c(\mathbf{x}_c^{(1)}, x_c^{(2)})$ defined over a two level hierarchical network with the addition of a single auxiliary variable $x_c^{(2)}$ for every clique $c \in \mathcal{C}$. This auxiliary variable take values from an extended label set $\mathcal{L}^e = \mathcal{L} \cup \{L_F\}$, where $L_F$, the 'free' label of the auxiliary variables, allows its child variables to take any label without paying a pairwise penalty.

In general, every higher order cost function can be converted to a 2−layer associative hierarchical network by taking an approach analogous to that of factor graphs (Kschischang et al., 2001) and adding a single multi-state auxiliary variable. However, to do this for general higher order functions requires the addition of an auxiliary variable with an exponential sized label set (Wainwright and Jordan, 2008). Fortunately, the class of higher order potentials we are concerned with can be compactly described as AHNs with auxiliary variables that take a similar sized label set to the base layer, permitting fast inference.

The corresponding higher order function can be written as:

$$\begin{aligned}
\psi_c(\mathbf{x}_c^{(1)}) &= \min_{x_c^{(2)}} \psi_c(\mathbf{x}_c^{(1)}, x_c^{(2)}) \\
&= \min_{x_c^{(2)}} \left[ \phi_c(x_c^{(2)}) + \sum_{i \in c} \phi_{ic}(x_i^{(1)}, x_c^{(2)}) \right] (5)
\end{aligned}$$

The unary potentials $\phi_c(x_c^{(2)})$ defined on the auxiliary variable $x_c^{(2)}$ assign the cost $\gamma_l$ if $x_c^{(2)} = l \in \mathcal{L}$, and $\gamma_{max}$ if $x_c^{(2)} = L_F$. The pairwise potential $\phi_{ic}(x_i, x_c^{(2)})$ is defined as:

$$\phi_{ic}(x_i, x_c^{(2)}) = \begin{cases} 0 & \text{if } x_c^{(2)} = L_F, \text{ or } x_c^{(2)} = x_i. \\ k_c^i & \text{if } x_c^{(2)} = l \in \mathcal{L}, \text{ and } x_i \neq l. \end{cases} \quad (6)$$

**General Formulation** The scheme described above can be extended by allowing pairwise and higher order potentials to be defined over $\mathbf{x}^{(2)}$ and further over $\mathbf{x}^{(i)}$, which corresponds to higher order potentials defined over the layer $\mathbf{x}^{(i-1)}$. The higher order energy corresponding to the general hierarchical network can be written using the following recursive function:

$$\begin{aligned}
E^{(1)}(\mathbf{x^{(1)}}) &= \sum_{i \in \mathcal{V}} \psi_i^{(1)}(x_i^{(1)}) + \sum_{ij \in \mathcal{E}^{(1)}} \psi_{ij}^{(1)}(x_i^{(1)}, x_j^{(1)}) \\
&+ \min_{\mathbf{x}^{(2)}} E^{(2)}(\mathbf{x}^{(1)}, \mathbf{x}^{(2)}) \quad (7)
\end{aligned}$$

where $E^{(2)}(\mathbf{x}^{(1)}, \mathbf{x}^{(2)})$ is recursively defined as:

$$\begin{aligned}
&E^{(n)}(\mathbf{x}^{(n-1)}, \mathbf{x}^{(n)}) \\
&= \sum_{c \in V^{(n)}} \phi_c(x_c^{(n)}) + \sum_{c \in \mathcal{V}^{(n)} i \in c} \phi_{ic}^{(n)}(x_i^{(n-1)}, x_c^{(n)}) \\
&+ \sum_{cd \in \mathcal{E}^{(n)}} \psi_{cd}^{(n)}(x_c^{(n)}, x_d^{(n)}) + \min_{\mathbf{x}^{(n+1)}} E^{(n+1)}(\mathbf{x}^{(n)}, \mathbf{x}^{(n+1)})
\end{aligned} \quad (8)$$

and $\mathbf{x}^{(n)} = \{x_c^{(n)} | c \in \mathcal{V}^n\}$ denotes the set of variables at the $n^{\text{th}}$ level of the hierarchy, $\mathcal{E}^{(n)}$ represents the edges at this layer, and $\psi_{ic}^{(n)}(\mathbf{x}_c^{(n-1)}, x_c^{(n)})$ denotes the inter-layer potentials defined over variables of layer $n-1$ and $n$.

While the hierarchical formulation of both Taskar's and Kohli's models can be understood as a mathematical convenience that allows for fast and efficient bounded inference, our earlier work (Ladicky et al., 2009) used it for true multi-scale inference, modelling constraints defined over many quantisations of the image.

## 4 INFERENCE

**Inference in Pairwise Networks** Although the problem of MAP inference is NP-hard for most associative pairwise functions defined over more than two labels, in real world problems many conventional algorithms provide near optimal solutions over grid connected networks (Szeliski et al., 2006). However, the dense structure of hierarchical networks results in frustrated cycles and makes traditional reparameterisation based message passing algorithms for MAP inference such as loopy belief propagation (Weiss and Freeman, 2001) and tree-reweighted message passing (Kolmogorov, 2006) slow to converge and unsuitable (Kolmogorov and Rother, 2006). Many of these frustrated cycles can be eliminated via the use of cycle inequalities (Sontag et al., 2008; Werner, 2009), but only by significantly increasing the run time of the algorithm. Graph cut based move making algorithms do not suffer from this problem and have been successfully used for minimising pairwise functions defined over densely connected networks encountered in vision.

Examples of move making algorithms include $\alpha$-expansion which can only be applied to metrics, $\alpha\beta$ swap which can be applied to semi-metrics (Boykov et al., 2001), and range moves (Kumar and Torr, 2008; Veksler, 2007) for truncated convex potentials. These moves differ in the size of the space searched for the optimal move. While expansion and swap search a space of size at most $2^n$ while minimising a function of $n$ variables, the range moves explores a much larger space of $K^n$ where $K$ is a parameter of the energy (see Veksler (2007) for more details). Of these move making approaches, only $\alpha\beta$ swap can be directly applied to associative hierarchical networks as the term $\phi_{ic}(x_i, x_c)$, is not a metric nor truncated convex.

These methods start from an arbitrary initial solution of the problem and proceed by making a series of changes each of which leads to a solution of the same or lower energy (Boykov et al., 2001). At each step, the algorithms project a set of candidate moves into a Boolean space, along with their energy function. If the resulting projected energy function (also called the *move energy*) is both submodular and pairwise, it can be exactly minimised in polynomial time by solving an equivalent st-mincut problem. These optima can then be mapped back into the original space, returning the optimal move within the move set. The move algorithms run this procedure until convergence, iteratively picking the best candidate as different choices of range are cycled through.

**Minimising Higher Order Functions** A number of researchers have worked on the problem of MAP inference in higher order AMNs. Lan et al. (2006) proposed approximation methods for BP to make efficient inference possible in higher order MRFs. This was followed by the recent works of Potetz and Lee (2008); Tarlow et al. (2008, 2010) in which they showed how belief propagation can be efficiently performed in networks containing moderately large cliques. However, as these methods were based on BP, they were quite slow and took minutes or hours to converge, and lack bounds.

To perform inference in the $P^n$ models, Kohli et al. (2007, 2008), first showed that certain projection of the higher order $P^n$ model can be transformed into submodular pairwise functions containing auxiliary variables. This was used to formulate higher order expansion and swap move making algorithms The only existing work that addresses the problem of bounded higher order inference is (Gould et al., 2009) which showed how theoretical bounds could be derived given move making algorithms that proposed optimal moves by exactly solving some sub-problem. In application they used approximate moves which do not exactly solve the sub-problems proposed. Consequentially, the bounds they derive do not hold for the methods they propose. However, their analysis can be applied to the $P^n$ (Kohli et al., 2007) model and inference techniques, which do propose optimal moves, and it is against these bounds that we compare our results.

### 4.1 INFERENCE WITH $\alpha$-EXPANSION

We show that by restricting the form of the inter-layer potentials $\psi_c^{(n)}(\mathbf{x}_c^{(n-1)}, x_c^{(n)})$ to that of the weighted Robust $P^n$ model (Kohli et al., 2008) (see (4)), we can apply $\alpha$-expansion to the pairwise form of the AHN.

This requires a transform of all functions in the pairwise representation, so that they can be representable as a metric (Boykov et al., 2001). This transformation is non-standard and should be considered a contribution of this work.

We alter the form of the potentials in two ways. First, we assume that all variables in the hierarchy take values from the same label set $\mathcal{L}^e = \mathcal{L} \cup \{L_F\}$. Where this is not true — original variables $\mathbf{x}^{(1)}$ at the base of the hierarchy can not take label $L_F$ — we artificially augment the label set with the label $L_F$ and associate an infinite unary cost with it. Secondly, we make the inter-layer pairwise potentials *symmetric* by performing a local reparameterisation operation.

**Lemma 1.** *The inter-layer pairwise functions*

$$\phi_{ic}^{(n)}(x_i^{(n-1)}, x_c^{(n)}) = \begin{cases} 0 & \text{if } x_c^{(n)} = L_F \text{ or } x_c^{(n)} = x_i^{(n-1)} \\ k_c^i & \text{if } x_c^{(n)} = l \in \mathcal{L} \text{ and } x_i^{(n-1)} \neq l \end{cases} \quad (9)$$

*of (8) can be written as:*

$$\begin{aligned} \phi_{ic}^{(n)}(x_i^{(n-1)}, x_c^{(n)}) &= \psi_i^{(n-1)}(x_i^{(n-1)}) + \psi_c^{(n)}(x_c^{(n)}) \\ &\quad + \Phi_{ic}^{(n)}(x_i^{(n-1)}, x_c^{(n)}), \end{aligned} \quad (10)$$

$$\text{where} \quad \Phi_{ic}^{(n)}(x_i^{(n-1)}, x_c^{(n)}) = \begin{cases} 0 & \text{if } x_i^{(n-1)} = x_c^{(n)} \\ k_c^i/2 & \text{if } x_i^{(n-1)} = L_F \\ & \text{or } x_c^{(n)} = L_F \\ & \text{and } x_i^{(n-1)} \neq x_c^{(n)} \\ k_c^i & \text{otherwise,} \end{cases} \quad (11)$$

*and*

$$\psi_c^{(n)}(x_c^{(n)}) = \begin{cases} 0 & \text{if } x_c^{(n)} \in \mathcal{L} \\ -k_c^i/2 & \text{otherwise,} \end{cases} \quad (12)$$

$$\psi_i^{(n-1)}(x_i^{(n-1)}) = \begin{cases} 0 & \text{if } x_i^{(n-1)} \in \mathcal{L} \\ k_c^i/2 & \text{otherwise.} \end{cases} \quad (13)$$

**Proof** *Consider a clique containing only one variable, the general case will follow by induction. Note that if*

*no variables take state $L_F$ the costs are invariant to reparameterisation. This leaves three cases:*

$$\begin{array}{|c|}
\hline
\mathbf{x_c^{(n)}} = \mathbf{L_F}, \mathbf{x_i^{(n-1)}} \in \mathcal{L} \\
\psi_c(x_c^{(n)}) + \psi_{ic}(x_c^{(n)}, x_i^{(n-1)}) = -k/2 + k/2 = 0 \\
\hline
\mathbf{x_c^{(n)}} \in \mathcal{L}, \mathbf{x_i^{(n-1)}} = \mathbf{L_F} \\
\psi_i(x_i^{(n-1)}) + \psi_{ic}(x_i^{(n-1)}, x_c^{(n)}) = k/2 + k/2 = k \\
\hline
\mathbf{x_c^{(n)}} = \mathbf{L_F}, \mathbf{x_i^{(n-1)}} = \mathbf{L_F} \\
\psi_i(x_i^{(n-1)}) + \psi_{ic}(x_i^{(n-1)}, x_c^{(n)}) + \psi_c(x_c^{(n)}) = \frac{k-k}{2} = 0 \\
\hline
\end{array}$$
(14)

**Bounded Higher Order Inference** We now prove bounds for $\alpha$-expansion over an AHN.

1. The pairwise function of lemma 1, is positive definite, symmetric, and satisfies the triangle inequality
$$\psi_{a,b}(x,z) \leq \psi_{a,b}(x,y) + \psi_{a,b}(y,z) \forall x,y,z \in \mathcal{L} \cup \{L_F\}. \quad (15)$$
Hence it is a metric, and the algorithms $\alpha\beta$ swap and $\alpha$-expansion can be used to minimise it.

2. By the work of Boykov et al. (2001), the $\alpha$-expansion algorithm is guaranteed to find a solution within a factor of $2\max\left(2, \max_{E \in \mathcal{E}^1} \frac{\max_{x_i,x_j \in \mathcal{L}} \psi_E(x_i,x_j)}{\min_{x_i,x_j \in \mathcal{L}} \psi_E(x_i,x_j)}\right)$ (i.e. 4 where the potentials defined over the base layer of hierarchy take the form of a Potts model) of the global optima.

3. The following two properties hold:
$$\min_{\mathbf{x}^{(1)}} E(\mathbf{x}^{(1)}) = \min_{\mathbf{x}^{(1)},\mathbf{x}^a} E'(\mathbf{x}^{(1)}) + E^a(\mathbf{x}^{(1)}, \mathbf{x}^a), \quad (16)$$
$$E(\mathbf{x}^{(1)}) \leq E'(\mathbf{x}^{(1)}) + E^a(\mathbf{x}^{(1)}, \mathbf{x}^a), \quad (17)$$
Hence, if there exists a labelling $(\mathbf{x}', \mathbf{x}^*))$ such that
$$E'(\mathbf{x}') + E^a(\mathbf{x}', \mathbf{x}^*) \leq k \min_{\mathbf{x}^{(1)},\mathbf{x}^a} E'(\mathbf{x}^{(1)}) + E^a(\mathbf{x}^{(1)}, \mathbf{x}^a). \quad (18)$$
then
$$E(\mathbf{x}') \leq k \min_{\mathbf{x}^{(1)}} E(\mathbf{x}^{(1)}). \quad (19)$$

Consequentially, the bound is preserved in the transformation that maps from the pairwise energy back to its higher order form. □

By way of comparison, the work of Gould et al. (2009) provides a bound of $2|c|$ for the higher order potentials of the strict $P^n$ model (Kohli et al., 2007), where $c$ is the largest clique in the network. Using their approach, no bounds are possible for the general class of Robust $P^n$ models or for associative hierarchical networks.

The moves of our new range-move algorithm (see next section) strictly contain those considered by $\alpha$-expansion and thus our approach automatically inherits the above approximation bound.

## 5 NOVEL MOVES AND TRANSFORMATIONAL OPTIMALITY

In this section we propose a novel graph cut based move making algorithm for minimising the hierarchical pairwise energy function defined in the previous section.

Let us consider a generalisation of the swap and expansion moves proposed in Boykov et al. (2001). In a standard swap move, the set of all moves considered is those in which a subset of the variables currently taking label $\alpha$ or $\beta$ change labels to either $\beta$ or $\alpha$. In our range-swap the moves considered allow any variables taking labels $\alpha, L_F$ or $\beta$ to change their state to any of $\alpha, L_F$ or $\beta$. Similarly, while a normal $\alpha$ expansion move allows any variable to change to some state $\alpha$, our range expansion allows any variable to change to states $\alpha$ or $L_F$.

This approach can be seen as a variant on the ordered range moves proposed in Veksler (2007); Kumar and Torr (2008), however while these works require that an ordering of the labels $\{l_1, l_2, \ldots, l_n\}$ exist such that moves over the range $\{l_i, l_{i+1} \ldots l_{i+j}\}$ are convex for some $j \geq 2$ and for all $0 < i \leq n - j$, our range moves function despite no such ordering existing.

We now show that the problem of finding the optimal swap move can be solved exactly in polynomial time. Consider a label mapping function $f_{\alpha,\beta} : \mathcal{L} \to \{1, 2, 3\}$ defined over the set $\{\alpha, L_F, \beta\}$ that maps $\alpha$ to 1, $L_F$ to 2 and $\beta$ to 3. Given this function, it is easy to see that the reparameterised inter-layer potential $\Phi_{ic}^{(n)}(x_i^{(n-1)}, x_c^{(n)})$ defined in lemma 1 can be written as a convex function of $f_{\alpha,\beta}(x_i^{(n-1)}) - f_{\alpha,\beta}(x_c^{(n)})$ over the range $\alpha, L_F, \beta$. Hence, we can use the Ishikawa construct (Ishikawa, 2003) to minimise the swap move energy to find the optimal move. A similar proof can be constructed for the range-expansion move described above.

The above defined move algorithm gives improved solutions for the hierarchical energy function used for formulating the object segmentation problem. We can improve further upon this algorithm. Our novel construction for computing the optimal moves explained in the following section, is based upon the original energy function (before reparameterisation) and has a strong transformational optimality property. We first describe the construction of a three label range move over the hierarchical network, and then show in section 5.2 that under a set of reasonable assumptions, our methods are equivalent to a swap or expansion move that exactly minimises the equivalent higher order energy defined over the base variables $E(\mathbf{x}^{(1)})$ of the hierarchical network (as defined in (7)).

## 5.1 CONSTRUCTION OF THE RANGE MOVE

We now explain the construction of the submodular quadratic pseudo boolean (QPB) move function for range expansion. The construction of the swap based move function can be derived from this range move.

In essence, we demonstrate that the cost function of (9) over the range $x_c \in \{\beta, L_F, \alpha\}, x_i \in \{\delta, L_F, \alpha\}$ where $\beta$ may or may not equal $\delta$ is expressible as a submodular QPB potential. To do this, we create a QPB function defined on 4 variables $c_1$, $c_2$, $i_1$ and $i_2$. We associate the states $i_1 = 1, i_2 = 1$ with $x_i$ taking state $\alpha$, $i_1 = 0, i_2 = 0$ with the current state of $x_i = \delta$, and $i_1 = 1, i_2 = 0$ with state $L_F$. We prohibit the state $i_1 = 0, i_2 = 1$ by incorporating the pairwise term $\infty(1 - i_1)i_2$ which assigns an infinite cost to the state $i_1 = 0, i_2 = 1$, and do the same respectively with $x_c$ and $c_1$ and $c_2$. To simplify the resulting equation, we write $I$ instead of $\Delta(\beta \neq \delta)$, $k_\delta$ for $\psi_{i,c}(L_F, \delta)$ and $k_\alpha$ for $\psi_{i,c}(L_F, \alpha)$ then

$$\psi_{i,c}(x_i, x_c) = (1-I)k_\delta c_2(1-x_2) - Ik_\delta c_2 + k_\alpha(1-c_1)x_1 \quad (20)$$

over the range $x_c \in \{\beta, L_F, \alpha\}, x_i \in \{\delta, L_F, \alpha\}$.

The proof follows from inspection of the function. Note that $c_2 = 1$ if and only if $x_c = \beta$ while $c_1 = 0$ if and only if $c = \alpha$. If $x_c = L_F$ then $c_2 = 0$ and $c_1 = 1$ and the cost is always 0. If $x_c = \alpha$ the first two terms take cost 0, and the third term has a cost of $k_\alpha$ associated with it unless $x_i = \alpha$. Similarly, if $x_c = \beta$ there is a cost of $k_\beta$ associated with it, unless $x_i$ also takes label $\beta$. □

## 5.2 OPTIMALITY

Note that both variants of unordered range moves are guaranteed to find the global optima if the label space of $\mathbf{x}^{(1)}$ contains only two states. This is not the case for the standard forms of $\alpha$ expansion or $\alpha\beta$ swap as auxiliary variables may take one of three states.

**Transformational optimality** Consider an energy function defined over the variables $\mathbf{x} = \{\mathbf{x}^{(h)}, h \in \{1, 2, \ldots, H\}\}$ of a hierarchy with $H$ levels. We call a move making algorithm *transformationally optimal* if and only if any proposed move $\mathbf{x}_* = \{\mathbf{x}_*^{(h)}, h \in \{1, 2, \ldots, H\}\}$ satisfies the property:

$$E(\mathbf{x}_*) = \min_{\mathbf{x}_*^{\text{aux}}} E(\mathbf{x}_*^{(1)}, \mathbf{x}_*^{\text{aux}}) \quad (21)$$

where $\mathbf{x}_*^{\text{aux}} = \bigcup_{h \in 2,\ldots,H} \mathbf{x}_*^{(h)}$ represents the labelling of all auxiliary variables in the hierarchy. Note that any move proposed by transformationally optimal algorithms minimises the original higher order energy (7). We now show that when applied to hierarchical networks, the *range* moves are transformationally optimal.

**Move Optimality** To guarantee transformational optimality we need to constrain the set of higher order potentials. Consider a clique $c$ with an associated auxiliary variable $x_c^{(i)}$. Let $\mathbf{x}_l$ be a labelling such that $x_c^{(i)} = l \in \mathcal{L}$ and $\mathbf{x}_{L_F}$ be a labelling that only differs from it that the variable $x_c^{(i)}$ takes label $L_F$. We say a clique potential is *hierarchically consistent* if it satisfies the constraint:

$$E(\mathbf{x}_l) \geq E(\mathbf{x}_{L_F}) \implies \frac{\sum_{i \in c} k_c^i \Delta(x_i = l)}{\sum_{i \in c} k_c^i} > 0.5. \quad (22)$$

The property of hierarchical consistency is also required in computer vision for the cost associated with the hierarchy to remain meaningful. The labelling of an auxiliary variable within the hierarchy should be reflected in the state of the clique associated with it. If an energy is not hierarchically consistent, it is possible that the optimal labelling of regions of the hierarchy will not reflect the labelling of the base layer.

The constraint (22) is enforced by construction, weighting the relative magnitude of $\psi_i(l)$ and $\psi_{i,j}(b_j, x_c^{(i)})$ to guarantee that:

$$\psi_i(l) + \sum_{j \in N_i/c} \max_{x_j \in \mathcal{L} \cup \{L_f\}} \psi_{i,j}(b_j, x_c^{(i)}) < 0.5 \sum_{X_i \in c} k_i \forall l \in \mathcal{L}. \quad (23)$$

If this holds, in the degenerate case where there are only two levels in the hierarchy, and no pairwise connections between the auxiliary variables, our network is exactly equivalent to the $P^n$ model.

At most one $l \in \mathcal{L}$ at a time can satisfy (22), assuming the hierarchy is consistent. Given a labelling for the base layer of the hierarchy $\mathbf{x}^{(1)}$, an optimal labelling for an auxiliary variable in $\mathbf{x}^{(2)}$ associated with some clique must be one of two labels: $L_F$ and some $l \in \mathcal{L}$. By induction, the choice of labelling of any clique in $\mathbf{x}^{(j)}$ must also be a decision between at most two labels: $L_F$ and some $l \in \mathcal{L}$.

## 5.3 TRANSFORMATIONAL OPTIMALITY UNDER UNORDERED RANGE MOVES

**Swap range moves**

Swap based optimality requires an additional constraint to that of (22), namely that there are no pairwise connections between variables in the same level of the hierarchy, except in the base layer. From (6) if an auxiliary variable $x_c$ may take label $\gamma$ or $L_F$, and one of its children $x_i | i \in c$ take label $\delta$ or $L_F$, the cost associated with assigning label $\gamma$ or $L_F$ to $x_c$ is independent of the label of $x_i$ with respect to a given move.

Under a swap move, a clique currently taking label $\delta \notin \{\alpha, \beta\}$ will continue to do so. This follows from (4) as the cost associated with taking label $\delta$ is only dependent upon the weighted average of child variables taking state $\delta$, and this remains constant. Hence the only clique variables that may have a new optimal labelling under the swap are those currently taking state $\alpha, L_F$ or $\beta$, and these can only transform to one of the states $\alpha, L_F$ or $\beta$. As the range moves map exactly this set of transformations, the move proposed must be transformationally optimal, and consequently the best possible $\alpha\beta$ swap over the energy (7).

**Expansion Moves**

In the case of a range-expansion move, we can maintain transformational optimality while incorporating pairwise connections into the hierarchy — provided condition (22) holds, and the energy can be exactly represented in our submodular moves.

In order for this to be the case, the pairwise connections must be both convex over any range $\alpha, L_F, \beta$ and a metric. The only potentials that satisfy this are linear over the ordering $\alpha, L_F, \beta \, \forall \alpha, \beta$. Hence all pairwise connections must be of the form:

$$\psi_{i,j}(x_i, x_j) = \begin{cases} 0 & \text{if } x_i = x_j \\ \lambda/2 & \text{if } x_i = L_F \text{ or } x_j = L_F \text{ and } x_i \neq x_j \\ \lambda & \text{otherwise.} \end{cases}$$
(24)

where $\lambda \in \mathbb{R}_0^+$. By lemma 1, it can be readily seen that the connections in the hierarchical network are a constrained variant of this form.

A similar argument to that of the optimality of $\alpha\beta$ swap can be made for $\alpha$-expansion. As the label $\alpha$ is 'pushed' out across the base layer, the optimal labelling of some $x^{(n)}$ where $n \geq 2$ must either remain constant or transition to one of the labels $L_F$ or $\alpha$. Again, the range moves map exactly this set of transforms and the suggested move is both transformationally optimal, and the best expansion of label $\alpha$ over the higher order energy of (7).

## 6 EXPERIMENTS

We evaluate $\alpha$-expansion, $\alpha\beta$ swap, TRW-S, Belief Propagation, Iterated Conditional Modes, and both the expansion and swap based variants of our unordered range moves on the problem of object class segmentation over the MSRC data-set (Shotton et al., 2006), in which each pixel within an image must be assigned a label representing its class, such as grass, water, boat or cow. We express the problem as a three layer hierarchy. Each pixel is represented by a random variables of the base layer. The second layer is formed by performing multiple unsupervised segmentations over the image, and associating one auxiliary variable with each segment. The children of each of these variables in $x^{(2)}$ are the variables contained within the segment, and pairwise connections are formed between adjacent segments. The third layer is formed in the same manner as the second layer by clustering the image segments. Further details are given in Ladicky et al. (2009).

We tested each algorithm on 295 test images, with an average of 70,000 pixels/variables in the base layer and up to 30,000 variables in a clique, and ran them either until convergence, or for a maximum of 500 iterations. In the table in figure 1 we compare the final energies obtained by each algorithm, showing the number of times they achieved an energy lower than or equal to all other methods, the average difference $E(\text{method}) - E(\text{min})$ and average ratio $E(\text{method})/E(\text{min})$. Empirically, the message passing algorithms TRW-S and BP appear ill-suited to inference over these dense hierarchical networks. In comparison to the graph cut based move making algorithms, they had higher resulting energy, higher memory usage, and exhibited slower convergence.

While it may appear unreasonable to test message passing approaches on hierarchical energies when higher order formulations such as (Komodakis and Paragios, 2009; Potetz and Lee, 2008) exist, we note that for the simplest hierarchy that contains only one additional layer of nodes and no pairwise connections in this second layer, higher order and hierarchical message-passing approaches will be equivalent, as inference over the trees that represent higher order potentials is exact. Similar relative performance by message passing schemes was observed in these cases. Further, application of such approaches to the general form of (7) would require the computation of the exact min-marginals of $E^{(2)}$, a difficult problem in itself.

In all tested images both $\alpha$-expansion variants outper-

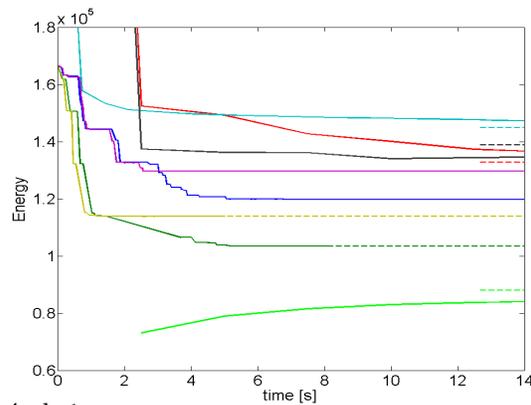
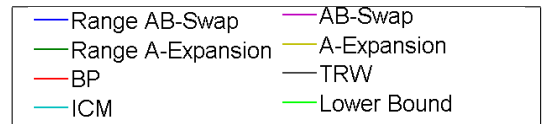

| Method | Best | $E(\text{meth}) - E(\text{min})$ | $\frac{E(\text{meth})}{E(\text{min})}$ | Time |
|---|---|---|---|---|
| Range-exp | 265 | 74.747887 | 1.000368 | 6.1s |
| Range-swap | 137 | 9033.847065 | 1.058777 | 19.8s |
| $\alpha$-expansion | 109 | 255.500278 | 1.001604 | 6.3s |
| $\alpha\beta$ swap | 42 | 9922.084163 | 1.060385 | 41.6s |
| TRW-S | 12 | 38549.214994 | 1.239831 | 8.3min |
| BP | 6 | 13455.569713 | 1.081627 | 2min |
| ICM | 5 | 45954.670836 | 1.277519 | 25.3s |

Obrázek 1: **Left** Typical behaviour of all methods along with the lower bound obtained from TRW-S an image from MSRC (Shotton et al., 2006) data set. The dashed lines at the right of the graph represent final converged solutions. **Right** Comparison of methods on 295 testing images. From left to right the columns show the number of times they achieved the best energy (including ties), the average difference ($E(\text{method}) - E(\text{min})$), the average ratio ($E(\text{method})/E(\text{min})$) and the average time taken. All three approaches proposed by this paper: $\alpha$-expansion under the reparameterisation of section 5, and the transformationally optimal range expansion and swap significantly outperformed existing inference methods both in speed and accuracy. See the supplementary materials for more examples.

formed TRW-S, BP and ICM. These later methods only obtained minimal cost labellings in images in which the optimal solution found contained only one label i.e. they were entirely labelled as grass or water. The comparison also shows that unordered range move variants usually outperform vanilla move making algorithms. The higher number of minimal labellings found by the range-move variant of $\alpha\beta$ swap in comparison to those of vanilla $\alpha$-expansion can be explained by the large number of images in which two labels strongly dominate, as unlike standard $\alpha$-expansion both range move algorithms are guaranteed to find the global optima of such a two label sub-problem (see section 5.2). The typical behaviour of all methods alongside the lower bound of TRW-S can be seen in figure 1 and further, alongside qualitative results, in the supplementary materials.

# 7 CONCLUSION

This paper shows that higher order AMNs are intimately related to pairwise hierarchical networks. This observation allowed us to characterise higher order potentials which can be solved under a novel reparameterisation using conventional move making expansion and swap algorithms, and derive bounds for such approaches. We also gave a new transformationally optimal family of algorithms for performing efficient inference in higher order AMN that inherits such bounds.

We have demonstrated the usefulness of our algorithms on the problem of object class segmentation where they have been shown to outperform state of the art approaches over challenging data sets (Ladicky et al., 2009) both in speed and accuracy.